\pdfoutput=1  
\documentclass[letterpaper, 10 pt, conference]{ieeeconf}  

\IEEEoverridecommandlockouts 

\usepackage{times}
\usepackage{multicol}
\usepackage[bookmarks=true]{hyperref}
\usepackage{graphicx}
\usepackage{subcaption}

\usepackage{tabularx}
\usepackage{booktabs}

\usepackage{amsmath}
\usepackage{amssymb}
\usepackage{bm}

\usepackage{siunitx}
\usepackage{microtype}

\usepackage[
	backend=biber, 
	bibencoding=utf8,
	style=ieee, 
	sorting=none, 
	natbib=true, 
	doi=false, 
	isbn=false, 
	url=false, 
	eprint=false, 
	date=year,
	maxcitenames=1,
	mincitenames=1,
	maxnames=2,
	minnames=1,
	dashed=false
]{biblatex}

\usepackage{xspace}

\usepackage[printonlyused]{acronym}
\acrodef{ICP}{Iterative Closest Point}
\acrodef{GNSS}{Global Navigation Satellite System}
\acrodef{HMI}{Hazardously Misleading Information}

\usepackage{lipsum}
\usepackage[dvipsnames]{xcolor}
\usepackage{algorithm}
\usepackage{algpseudocode}

\newcommand{\norm}[1]{\left\lVert#1\right\rVert}

\newcommand{\varPref}{{}^{w}\mathcal{P}_{\mathrm{ref}}}
\newcommand{\varTref}{{}^{w}\mathcal{T}_{\mathrm{ref}}}

\newcommand{\varCFrame}{\mathcal{W}}
\newcommand{\varWheelFP}{\mathcal{F}}
\newcommand{\varHB}{{}^{w}\mathcal{H}_\mathrm{b}}
\newcommand{\varHF}{{}^{w}\mathcal{H}_\mathrm{f}}
\newcommand{\varHDelta}{{}^{w}\mathcal{H}_\Delta}
\newcommand{\varHSim}{{}^{w}\mathcal{H}_\mathrm{sim}}
\newcommand{\varPb}[1][]{{}^{w}\mathcal{P}_\mathrm{b#1}}
\newcommand{\varPf}[1][]{{}^{w}\mathcal{P}_\mathrm{f#1}}
\newcommand{\varPbe}[1][]{{}^{w}\mathcal{P}_\mathrm{be#1}}
\newcommand{\varPfe}[1][]{{}^{w}\mathcal{P}_\mathrm{fe#1}}
\newcommand{\varLf}[1][]{{}^f\mathcal{L}_{#1}}
\newcommand{\varLb}[1][]{{}^b\mathcal{L}_{#1}}

\newcommand{\varTextCalib}{{}^{f}\mathbf{T}_{b}}
\newcommand{\varPathWidth}{d_w}

\newcommand{\varCompRatio}{\raisebox{0.15em}{$\chi$}}
\newcommand{\varAngleRepose}{\beta}

\newcommand{\varSink}{d}
\newcommand{\varSinkOne}{d_1}  
\newcommand{\varSinkTwo}{d_2}  
\newcommand{\varFlux}{\boldsymbol{\varphi}}
\newcommand{\varParamVec}{\boldsymbol{\theta}}
\newcommand{\varSim}{\mathcal{S}}
\newcommand{\varCost}{c}
\newcommand{\varWinCells}{\Omega}  
\newcommand{\varWinLen}{W}  
\newcommand{\varVin}{\mathcal{V}}
\newcommand{\varVc}{\mathcal{V}_{\mathrm{c}}}
\newcommand{\varVd}{\mathcal{V}_{\mathrm{d}}}

\AtEveryBibitem{%
  \clearfield{isbn}%
  \clearfield{pagetotal}%
  \clearfield{issuetitle}%
  \clearfield{series}%
  \clearfield{note}%
  \clearfield{eventtitle}%
  \clearlist{location}%
}

\acrodef{MPC}{Model Predictive Control}
\acrodef{DEM}{Discrete Element Method}
\acrodef{FEM}{Finite Element Method}
\acrodef{SCM}{Soil Contact Modeling}
\acrodef{ROI}{Region Of Interest}

\def\authorrefmark#1{\ensuremath{^{\textbf{#1}}}}

\title{\LARGE \bf Tracking the Ground: Online Lidar Identification of Robot-Induced Soil Deformation in Agricultural Environments}

\author{Tom Montagnon\authorrefmark{1}, Johann Laconte\authorrefmark{1},
	Benoit Thuilot\authorrefmark{2}, Wonjae Cho\authorrefmark{3},
	Roland Lenain\authorrefmark{1}%
\thanks{\authorrefmark{1}Université Clermont Auvergne, INRAE, UR TSCF, 63000,
	Clermont-Ferrand, France}%
\thanks{\authorrefmark{2}Institut Pascal, Université Clermont Auvergne, Clermont Auvergne INP,
	CNRS, 63000 Clermont-Ferrand, France}%
\thanks{\authorrefmark{3}National Agriculture and Food Research Organization, 1-31-1 Kannondai,
	Tsukuba, Ibaraki 305-0856, Japan}%
}

\begin{document}

\maketitle
\thispagestyle{empty}
\pagestyle{empty}

\begin{abstract}
	Agriculture faces many challenges, and robotic systems can play an important role in addressing them by improving the efficiency and sustainability of field operations.
	Among these challenges, preserving soil health is a critical concern, as vehicle-soil interactions can degrade the soil structure and produce unwanted surface deformation.
	A key step toward soil-aware robotics is to explicitly account for how vehicle traffic deforms the ground, yet soil state is typically not treated as a variable.
	We address this gap by proposing a framework to quantify traffic-induced soil deformation and estimate its evolution online from lidar observations.
	The method relies on a reduced-order parametric model that represents the soil behavior via physically interpretable parameters, yielding a continuously updated and observable representation of soil state.
	Experiments conducted in different soil conditions demonstrate the ability of the approach to capture deformation induced by the robot.
	By making soil response measurable and interpretable during operation, the proposed framework establishes a basis for soil-aware robotic operation, in which the estimated state can be exploited to adapt robotic behaviors in order to reduce soil degradation.
\end{abstract}

\section{INTRODUCTION}

Off-road mobile robots are becoming increasingly popular due to their potential societal impact, in domains such as agriculture, space, and exploration.
In agriculture specifically, meeting the food demand of a growing population has long driven the search for higher productivity, accompanied by structural changes such as farm consolidation and larger, more powerful machinery \cite{macdonald_three_2018}.
These changes have also intensified concerns about the long-term sustainability of soil use \cite{lal_restoring_2015}.

\begin{figure}[t]
	\centering
	\includegraphics[width=\columnwidth]{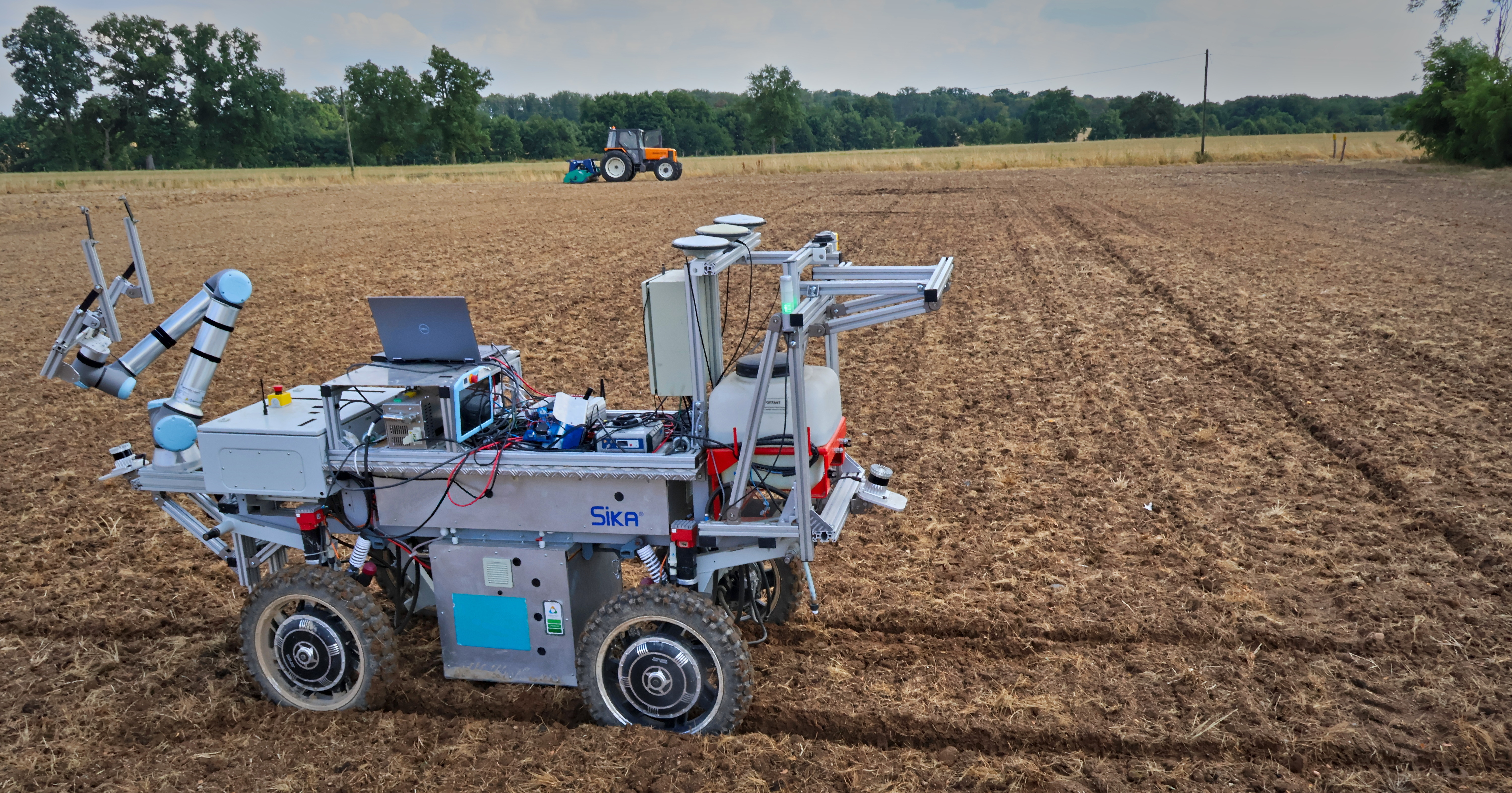}
	\caption{Illustration of a rut formed by the robotic platform after
		two passes on freshly watered soil, highlighting the cumulative
		soil degradation that can result from repeated vehicle traffic.}
	\label{fig:rut}
\end{figure}

In response, agroecology has gained attention as a framework reconciling productivity with environmental preservation through locally adapted, resource-efficient practices \cite{wezel_agroecological_2020}.
Soil is not merely a support for vehicle motion but a living, structured medium conditioning water transfer, root development, and crop functioning.
Repeated vehicle traffic induces compaction, rutting, and structural degradation, as illustrated in \autoref{fig:rut}, ultimately decreasing productivity and making soil preservation a central requirement for sustainable field operations \cite{nawaz_soil_2013}.

However, replacing heavy tractors with lighter robots does not naturally solve the problem: even a lightweight autonomous robot can significantly alter topsoil properties, since the issue lies not only in vehicle mass or wheel properties but in the cumulative effect of trajectory repetition on the soil, depending on its conditions \cite{calleja-huerta_impacts_2023}.
This motivates a shift toward robot designs that explicitly account for soil preservation during operation.

This work contributes to such a shift by estimating soil state as an explicit variable, allowing robot behavior to be adapted to limit its degradation.
Specifically, this paper proposes a framework that measures traffic-induced soil deformation from lidar data and identifies, online, a compact model of soil response directly from field observations.
This model supports continuous, physically grounded monitoring of soil state.
Once quantified, soil deformation can be used to adapt robot trajectories, manage energy consumption, or adjust traction strategies as terrain conditions degrade.
Each use case requires the same prerequisite: soil parameters that are physically meaningful, identifiable online from lidar data alone, and predictive of the deformation they anticipate.
This paper focuses on establishing this prerequisite, leaving its exploitation by a downstream controller to future work.
Its main contributions are as follows:
\begin{itemize}
	\setlength\itemsep{1pt}
	\setlength\parskip{0pt}
	\setlength\topsep{1pt}

	\item A lidar-based processing pipeline that quantifies soil deformation
	      induced by robot passes through differential heightmap analysis;
	\item A reduced-order parametric model of soil deformation, with three
	      physically interpretable parameters, identified online from
	      sequential lidar data; and
	\item An experimental validation demonstrating the applicability of the
	      approach across diverse soil conditions.
\end{itemize}

\section{RELATED WORK}

Soil compaction caused by field traffic is a well-established limitation of mechanized agriculture: repeated loading alters pore structure, restricts root development, and degrades soil functions \cite{batey_soil_2009}, and the historical increase in machinery mass has only deepened this concern by amplifying stress transmission into the soil profile \cite{keller_historical_2019}.
The dominant response has been Controlled Traffic Farming, which confines wheel passages to permanent lanes and has proven effective at limiting compaction for conventional machinery \cite{chamen_controlled_2015}.

Lightweight autonomous robots, however, change the terms of the problem rather than removing it: recent experimental work shows that even a light robotic platform can induce significant topsoil degradation when the same track is repeatedly loaded \cite{calleja-huerta_impacts_2023}, and that structural alterations may persist through seasons when operations are carried out on weak soil \cite{calleja-huerta_evolution_2024}.
The question for robotic systems is therefore not only whether a vehicle has a good configuration (e.g., mass, contact surface), but how its trajectory and repetition pattern translate into cumulative soil damage.
Accounting for this cumulative damage has motivated a parallel line of work on soil-aware traffic planning.
Early work in this direction relied on static susceptibility maps as a proxy for the state of the soil: routing decisions were adapted to spatial soil variability to reduce compaction risk \cite{bochtis_dss_2012}, extended to erosion on sloped terrain \cite{spekken_planning_2016}, and applied to vineyard path planning under compaction constraints \cite{santos_path_2018}.

More recent approaches have moved toward explicit traffic-impact models.
Soil2Cover integrates a compaction cost function directly into a coverage path planner, jointly optimizing route efficiency and predicted soil damage \cite{mier_soil2cover_2025}; similarly, \textcite{calleja-huerta_new_2026} propose a spatial framework that links wheel-track localization, traffic intensity, and empirical soil response to estimate the in-field distribution of compaction.
These contributions make soil-aware planning more operational, but they still rely on mapped proxies or empirical update laws rather than on direct observation of robot-induced deformation.

This gap is further accentuated by the practical difficulty of acquiring dense soil-state information in the field \cite{alaoui_mapping_2018}.
Robot-assisted probing can build such a map by combining autonomous exploration with kriging-based compaction mapping \cite{fentanes_3-d_2018}.
Every sample requires halting to drive a penetrometer into the ground, so mapping a field is a slow survey in its own right and cannot be run during a productive field operation.
Non-contact sensing avoids this cost: \textcite{campbell_modeling_2013} showed that lidar-derived surface geometry can support the estimation of rutting-related soil response.
However, no existing framework turns such geometric observations into a compact, identifiable model of robot-induced terrain deformation.
Building such a model in turn requires relating wheel loading to soil deformation in the first place, a relationship long studied in classical terramechanics.
The pressure-sinkage relationship introduced by \textcite{bekker_theory_1956} remains the standard starting point for relating normal loading to soil deformation, and the synthesis by \textcite{wong_terramechanics_2010} turned these local laws into a broader engineering framework for off-road mobility.
Modern reviews confirm that such semi-empirical models remain attractive for real-time applications, while also pointing to their main weakness: the need for offline terrain parameter calibration and their simplified treatment of soil behavior \cite{he_review_2019}.

Several works have extended this foundation toward explicit terrain deformation in simulation.
\textcite{krenn_simulation_2008} embedded a deformable-terrain interaction model into a rover locomotion simulator; \textcite{buse_scm_2016} developed \ac{SCM} as a modular soil contact model coupling terramechanics laws with height-field deformation; and \textcite{holz_soil_2009} proposed a hybrid approach combining a global surface representation with local particle-based behavior near the contact zone.
Beyond agriculture, related challenges arise in heavy off-road contexts, where \textcite{wiberg_discrete_2021} studied large deformations under heavy vehicles with \ac{DEM}, and \textcite{edwin_soft_2018} proposed a fast soft-soil track model for tracked vehicles.
Across these domains, the recurring tension between physical fidelity and computational tractability mirrors the constraints that arise in field robotics.

Adjacent earthmoving and off-road domains follow similar directions: \textcite{miron_towards_2022} place a heightmap of the worksite in the planning loop for autonomous grading, and \textcite{jaiswal_deformable_2019} build a real-time deformable-terrain simulation of a tractor with a front-loader.
Both, however, run a forward model with predefined terrain parameters rather than identifying soil properties from field data.

\begin{figure*}
	\centering

	\resizebox{\textwidth}{!}{
		\includegraphics{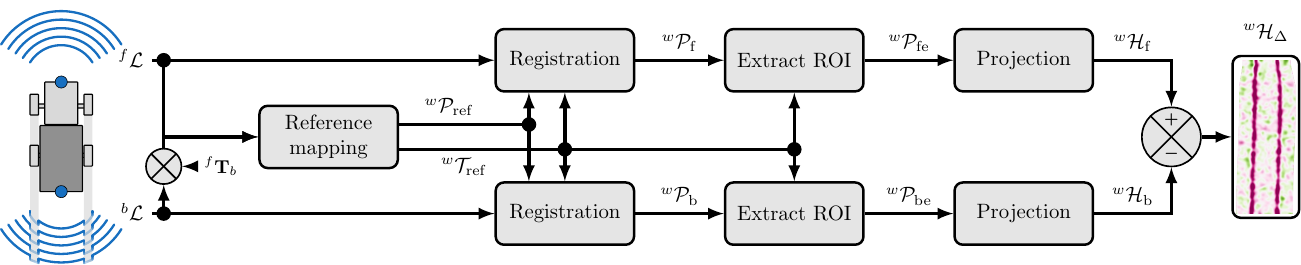}
	}
	\caption{
		Overview of the proposed processing pipeline.
		Raw lidar scans from the front sensor $\varLf$ and back sensor $\varLb$ are first fused using the extrinsic calibration $\varTextCalib$; the resulting merged point cloud is processed by the reference mapping block (via \ac{ICP}), which produces a reference map $\varPref$ and a reference trajectory $\varTref$.
		Both individual sensor streams are then independently registered (via \ac{ICP}) onto the reference map, yielding globally aligned point clouds $\varPf$ and $\varPb$.
		The "Extract ROI" blocks spatially filter each aligned cloud along the reference trajectory, producing $\varPfe$ and $\varPbe$.
		The filtered clouds are subsequently projected onto regular heightmaps $\varHF$ and $\varHB$, whose cell-wise difference yields the final displacement map $\varHDelta$.
	}
	\label{fig:pipeline}
\end{figure*}

Unlike these approaches, we do not assume that terrain parameters are known in advance: instead, we estimate a reduced-order soil representation directly from lidar observations of robot-induced surface change, using the pipeline shown in \autoref{fig:pipeline}.

\section{DATA ACQUISITION AND PROCESSING} \label{sec:pipeline}

This section describes the pipeline used to measure robot-induced soil deformation from lidar data, illustrated in \autoref{fig:pipeline}.
The proposed approach relies on a dual-lidar configuration: a front sensor acquires the terrain before wheel contact, and a back sensor captures the same area afterward, so that their difference directly observes the deformation caused by the vehicle.
Quantities below are expressed in one of three coordinate frames: the front sensor frame~$f$, the back sensor frame~$b$, or a fixed world frame~$\varCFrame$, identified by a left superscript (e.g.\ $\varLf$, $\varLb$, ${}^w\mathcal{H}$).

Before a reference map can be built, both sensor streams must be expressed in a common frame.
Let $\varTextCalib \in \mathrm{SE}(3)$ denote the offline-estimated extrinsic calibration expressing the back sensor frame in the front sensor frame.
At each time step, the back scan is transformed and merged with the front scan:
\begin{equation}
	\varLf \cup \varTextCalib \cdot \varLb.
\end{equation}

The merged scans are fed to an \ac{ICP}-based mapper \cite{pomerleau_comparing_2013}, which incrementally registers successive scans into a sparse global map $\varPref$ and reference trajectory $\varTref$.
The map is kept sparse for computational tractability and robustness to small surface irregularities.
$\varTref$ mainly serves as an initial guess for later registrations.

The incremental mapper registers each scan against its predecessors, so its pose estimate drifts along the path, by different amounts for the front and back streams.
Differencing the two heightmaps directly would then capture this drift as much as the deformation.
We instead re-register the full front and back point clouds, by \ac{ICP}, onto the sparse reference map, a single globally consistent frame, each seeded by the temporally closest pose of $\varTref$.
This step brings the dense front and back data into the common frame the mapper established, so that any residual error the reference carries is identical for both maps and cancels in $\varHB - \varHF$.
The two resulting dense maps $\varPf$ and $\varPb$ therefore differ only where the ground actually changed.

We then spatially filter these dense maps, keeping only points of the \ac{ROI} corresponding to a corridor of width $\varPathWidth$ around the trajectory.
The filtered clouds are rasterized onto a shared regular grid in the horizontal $(x,y)$ plane of $\varCFrame$, with cell $(i,j)$ of edge length $\Delta x$, covering the extracted corridor.
Each cell's elevation is the average $z$-coordinate of the points it contains, with empty cells filled by local interpolation.
As such, this produces two heightmaps $\varHF$ and $\varHB$ corresponding to the state of the ground before and after the traversal of the robot.
These two maps will be used to estimate the underlying parameters of the soil deformation model, as presented in \autoref{sec:model}.
Finally, the displacement map, \begin{equation} 	\varHDelta = \varHB - \varHF \end{equation} depicts the impact of the robot on the ground.

\section{SOIL DEFORMATION MODEL} \label{sec:model}

The overall goal of this section is to design a simulation starting from the initial heightmap $\varHF$ and estimate the state of the soil after the robot's traversal $\varHB$.
High-fidelity approaches such as \ac{FEM} and \ac{DEM} can in principle predict such deformation processes.
However, these approaches require substantial constitutive calibration.
They also remain too computationally demanding for online identification in field robotics \cite{he_review_2019}.
Classical terramechanics reduces this complexity by replacing full soil continuum modeling with compact parametric laws.
Yet these models usually rely on terrain parameters that must be identified from dedicated, invasive experiments and are difficult to obtain during field deployments \cite{he_review_2019}.

Rather than requiring such dedicated calibration, we propose a reduced-order model parameterized by three scalar variables: the sinkage depth~$\varSink$, the displacement-compaction ratio~$\varCompRatio$, and the angle of repose~$\varAngleRepose$.
Each parameter condenses several physical properties of the soil, such as stiffness, cohesion, frictional resistance, and initial structural state, into a single observable quantity that can be estimated directly from lidar data.
We first introduce this model from a theoretical standpoint, motivating each parameter in \autoref{sec:model:forward}, before detailing its numerical implementation in \autoref{subsec:implementation} and its online identification in \autoref{sec:model:estimation}.

\subsection{Parametric Soil Deformation Model}
\label{sec:model:forward}

\begin{figure}[h]
	\centering
	\includegraphics[width=\columnwidth]{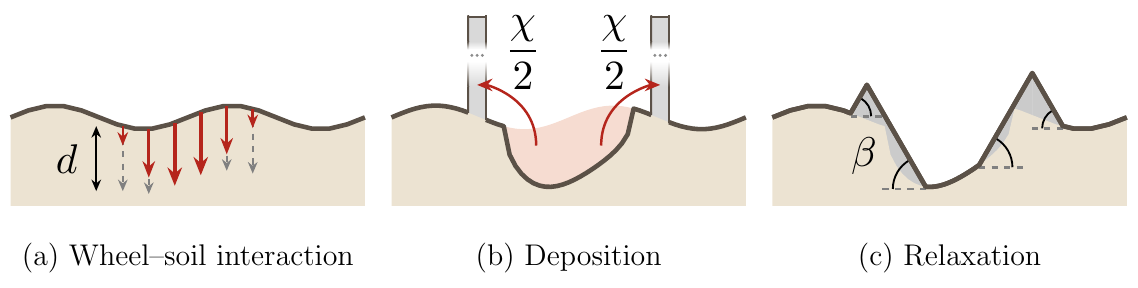}
	\caption{
		Schematic cross-section of the proposed soil deformation model.
		(a) The wheel sinks into the soil to a depth $\varSink$ (gray arrows), scaled per cell to the applied depth (red arrows).
		(b) A share $\varCompRatio$ of the removed soil is pushed to the side of the wheel track.
		(c) The soil then relaxes until no slope exceeds the angle of repose $\varAngleRepose$.
	}
	\label{fig:model_schematic}
\end{figure}

In this section, we detail each parameter, following the three stages illustrated in \autoref{fig:model_schematic}.
The interaction between the robot's wheel and the soil is simulated by first computing the displaced soil volume, then depositing a portion of it alongside the wheel track, and finally allowing the resulting soil heaps to relax toward a mechanically stable state.

\paragraph{Sinkage depth $\varSink$}

In classical terramechanics, the normal response of deformable soil is commonly described by Bekker's pressure-sinkage relationship \cite{wong_terramechanics_2010}:
\begin{equation}
	p = \left(\frac{k_c}{b} + k_\phi\right) \varSink^n,
	\label{eq:bekker}
\end{equation}

where $p$ is the applied normal pressure, $b$ is a characteristic contact width, $\varSink$ is the sinkage, $k_c$ and $k_\phi$ are respectively the cohesive and frictional moduli of terrain deformation, and $n$ is the (soil-dependent) sinkage exponent; together, these terrain parameters govern the compressive response of the soil.
Rearranging the equation gives the following expression of the sinkage depth:
\begin{equation}
	\varSink = \left(\frac{p}{\frac{k_c}{b} + k_\phi}\right)^{1/n}.
	\label{eq:sinkage}
\end{equation}
In practice, however, the parameters $k_c$, $k_\phi$, and $n$ are soil-dependent and are not directly available in field operation without dedicated identification procedures \cite{he_review_2019}.
For this reason, $\varSink$ is treated as a free parameter identified directly from the lidar height-map observations.
In this role, $\varSink$ serves as a compact descriptor of the effective normal response of the loaded soil, illustrated in \autoref{fig:model_schematic}(a).

\paragraph{Displacement-compaction ratio $\varCompRatio$}

The displaced soil is partitioned between compaction beneath the wheel and lateral displacement alongside the wheel track, with their relative contributions controlled by the displacement-compaction ratio.
We denote these two mechanisms by the volumetric compaction $\varVc$ and the lateral displacement $\varVd$, together accounting for the total volume $\varVin = \varVc + \varVd$ missing below the original surface after sinkage.
From this, we define the displacement-compaction ratio $\varCompRatio \in [0,1]$, illustrated in \autoref{fig:model_schematic}(b), as the proportion of soil being displaced:
\begin{equation}
	\varCompRatio = \frac{\varVd}{\varVd + \varVc}.
	\label{eq:chi_def}
\end{equation}
Accordingly, a value $\varCompRatio = 1$ corresponds to purely lateral displacement, while $\varCompRatio = 0$ indicates that the entire volume change is absorbed by volumetric compaction.
In practice, $\varCompRatio$ may exceed $1$.
A densely packed granular soil must loosen in order to shear, so its bulk volume increases as it is pushed aside.
This effect, known as dilatancy \cite{roquier_evaluation_2023}, makes the displaced volume larger than the volume removed by sinkage alone.

\paragraph{Surface relaxation and angle of repose $\varAngleRepose$}

The displaced material is deposited near the disturbed zone, producing local height discontinuities that must be smoothed to recover a realistic surface profile, illustrated in \autoref{fig:model_schematic}(c).
The evolution of the surface height $h(x,y,t)$ is governed by the Exner equation~\cite{paola_generalized_2005}:
\begin{equation}
	\frac{\partial h}{\partial t} = -\frac{1}{\varepsilon}\,\nabla \cdot \varFlux,
	\label{eq:exner}
\end{equation}
where $\varFlux\colon \mathbb{R}^2\to\mathbb{R}^2$ is the lateral soil flux, defined as
\begin{equation}
	\varFlux= -\max\left(\norm{\nabla h}-\tan\beta, 0\right)\cdot\frac{\nabla h}{\norm{\nabla h}}.
	\label{eq:flux}
\end{equation}
Material transfers only where the local slope exceeds the angle of repose $\varAngleRepose$, with an amplitude proportional to this slope excess and a direction along the steepest descent, as shown in \autoref{fig:model_schematic}(c).
Since only the steady-state profile is of interest, $\varepsilon$ merely sets the rate at which the surface relaxes and has no influence on the final, converged height field.
Its discretized form is detailed in \autoref{subsec:implementation}.

\subsection{Implementation of the model}
\label{subsec:implementation}

The terrain is discretized on a regular grid of resolution $\Delta x$, where $h_{i,j}$ denotes the elevation of cell $(i,j)$.
Simulation proceeds as summarized in \autoref{alg:sim}: for each pose along the trajectory, the wheel digs into the terrain and the removed material is redistributed onto the surrounding surface; once the full trajectory has been processed, the accumulated surface is relaxed to a stable configuration.
The remainder of this subsection details these three stages in the order they are executed.

\begin{algorithm}
	\caption{Forward soil deformation simulation $\varSim(\varHF,\,\varTref,\,\varWheelFP,\,\varParamVec)$}
	\label{alg:sim}
	\begin{algorithmic}[1]
		\Require \begin{tabular}[t]{@{}l@{}}
			$\varHF$: initial height map                                                                  \\
			$\varTref = \{(\mathbf{p}_t, \mathbf{R}_t)\}_{t=1}^N$: trajectory                             \\
			$\varWheelFP$: wheel footprint mask                                                           \\
			$\varParamVec = (\varSinkOne, \varSinkTwo, \varCompRatio, \varAngleRepose)$: model parameters \\
		\end{tabular}
		\State $\varHSim \leftarrow \varHF$
		\For{each pose $\mathbf{T} \in \varTref$, each wheel $\varWheelFP$}
		\State Apply the sinkage $\varSink_k$ over the wheel footprint \Comment{\eqref{eq:lower}}
		\State Compute the displaced volume $\varVin$ \Comment{\eqref{eq:total_volume}}
		\State Deposit $\varCompRatio \cdot \varVin$ sideways of the wheel
		\EndFor
		\Repeat
		\State Compute the flux $\varFlux_{i,j}(\varAngleRepose)$ for all cells \Comment{\eqref{eq:flux}}
		\State Update the height map $\varHSim$ \Comment{\eqref{eq:height_update}}
		\Until{convergence}
		\State \Return $\varHSim$
	\end{algorithmic}
\end{algorithm}

\subsubsection{Wheel-soil interaction}

Each wheel has a circular footprint $\varWheelFP$ whose diameter matches the wheel width, illustrated in \autoref{fig:model_schematic}(a).
In practice, one instance of $\varSink$ is identified per wheel track: the two-axle platform considered here produces two distinct ruts, one per side, hence the two sinkage depths $\varSinkOne$, $\varSinkTwo$.
This generalizes naturally to more complex configurations.
For instance, a vehicle with independently steered front and rear axles, whose front and rear wheels then follow separate tracks, would cut four distinct ruts and require four independent sinkage depths.
At each pose, the interaction proceeds in two steps.

First, for every cell $(i,j)$ of the footprint, a local reference elevation $\tilde{h}_{i,j}$ is computed by averaging the terrain over a small neighborhood around it, below radius half the wheel width.
This lets the reference follow the real topography under the wheel, rather than imprinting the circular outline of $\varWheelFP$ into the terrain.
It also smooths out high-frequency topography that would naturally be smoothed out in reality by the wheel pressure.

Second, the cell is dug down by the nominal sinkage depth, scaled by a centrality factor $s_{i,j} \in [0,1]$ that is equal to $1$ along the wheel's centerline and tapers smoothly to $0$ at the lateral edge of $\varWheelFP$, to be adjusted depending on the wheel shape:
\begin{equation}
	h_{i,j} \leftarrow \min\!\left(h_{i,j},\;\tilde{h}_{i,j} - s_{i,j}\,\varSink_k\right), \quad \forall\,(i,j)\in\varWheelFP,
	\label{eq:lower}
\end{equation}
with $k\in\{1,2\}$ selecting $\varSinkOne$ or $\varSinkTwo$ according to the side of the robot considered.
The minimum function filters out edge effects where driving over a small pre-existing hole would otherwise raise it back up.
Let $r_{i,j}$ denote the per-cell height removed by~\eqref{eq:lower}.
The corresponding quantity removed over the footprint,
\begin{equation}
	\varVin = \Delta x^2 \!\!\sum_{(i,j)\in\varWheelFP}\!\! r_{i,j},
	\label{eq:total_volume}
\end{equation}
is the same $\varVin$ defined in \autoref{sec:model:forward}.

\subsubsection{Material redistribution}

The removed volume $\varVin$ is redistributed, illustrated in \autoref{fig:model_schematic}(b), by depositing $\varCompRatio \cdot \varVin$ uniformly over the single-cell-wide band of cells adjacent to the wheel footprint, producing large local height discontinuities that the surface-relaxation step subsequently smooths out.

\subsubsection{Surface relaxation}

Once every pose has been processed, the accumulated height map is relaxed, illustrated in \autoref{fig:model_schematic}(c), by discretizing the Exner equation~\eqref{eq:exner}.
Decomposing the divergence along both axes and discretizing it as the net flux balance across each cell, the height at simulation step $n+1$ is updated as $h^{n+1}_{i,j} = h^n_{i,j} + \Delta h_{i,j}$, with
\begin{equation}
	\Delta h_{i,j}
	= -\frac{1}{\varepsilon\,\Delta x}\!\left(
	\varFlux^x_{i,j} - \varFlux^x_{i-1,j}
	+ \varFlux^y_{i,j} - \varFlux^y_{i,j-1}
	\right),
	\label{eq:height_update}
\end{equation}
where $\varFlux^x_{i,j}$ and $\varFlux^y_{i,j}$ are the components of $\varFlux_{i,j}$, taken as the flux leaving cell $(i,j)$ through its faces shared with $(i+1,j)$ and $(i,j+1)$.
At each cell, $\varFlux_{i,j}$ follows from \eqref{eq:flux} with the approximated gradient
\begin{equation}
	\nabla h_{i,j} \approx \frac{1}{\Delta x}[h_{i+1,j}-h_{i,j}, h_{i,j+1}-h_{i,j}]^T.
\end{equation}

The kernel iterates until the largest gradient norm $\norm{\nabla h_{i,j}}$ across all cells exceeds the angle-of-repose slope $\tan(\varAngleRepose)$ by less than a small tolerance $\tau$:
\begin{equation}
	\max_{i,j} \norm{\nabla h_{i,j}} - \tan(\varAngleRepose) < \tau.
	\label{eq:convergence}
\end{equation}
Once convergence is achieved, this yields the simulated heightmap $\varHSim$.
Note that this surface relaxation generalizes the approach of \textcite{yu_modeling_2024} by allowing flux along the local steepest-descent direction rather than restricting it to eight discrete directions.
Finally, this algorithm is easily parallelizable, since the flux \eqref{eq:flux} and subsequent update \eqref{eq:height_update} can be processed for all the cells at the same time.

\subsection{Parameter Estimation}
\label{sec:model:estimation}

Since soil properties may vary spatially along the path, the optimization is not performed on the full trajectory at once.
Instead, the trajectory is divided into short segments of length $\varWinLen$, each covering a compact spatial region, and parameters are estimated independently for each segment.

To identify the soil parameters $\varParamVec = (\varSink_1, \varSink_2, \varCompRatio, \varAngleRepose)$ over one such segment, we exploit the fact that the robot carries both a forward-looking and a backward-looking lidar.
As the robot traverses the terrain, the front sensor $\varLf$ acquires the heightmap $\varHF$ before any wheel contact, while the back sensor $\varLb$ measures the same zone after the passage, yielding $\varHB$.
We therefore run the forward simulation $\varSim$ of \autoref{alg:sim} on $\varHF$ to predict the post-passage surface $\varHSim$, and search for the parameters $\varParamVec$ that best reproduce $\varHB$.

\subsubsection{Cost Function and Gradient}

Letting $\varHSim(\varParamVec) = \varSim(\varHF,\varTref,\varWheelFP,\varParamVec)$ denote the simulated heightmap and $\varWinCells$ the set of grid cells within the current window's spatial footprint with a valid lidar measurement in both $\varHSim$ and $\varHB$ (excluding cells for which local interpolation could not fill a missing value for lack of nearby neighbors), the cost is the mean weighted L1 height difference against the lidar-measured back heightmap $\varHB$:
\begin{equation}
	\varCost(\varParamVec) = \frac{\sum_{(i,j)\in\varWinCells} w_{i,j}
		\left|\varHSim(\varParamVec)(i,j) - \varHB(i,j)\right|}
	{\sum_{(i,j)\in\varWinCells} w_{i,j}},
	\label{eq:cost}
\end{equation}
where $w_{i,j} \in \{0,1\}$ marks cells within about two wheel widths of the trajectory, restricting the cost to the zone of interest rather than diluting it over the full window.

As $\varHF$ and $\varTref$ fixed, $\varSim$ admits no analytical Jacobian, the gradient is approximated by forward finite differences.
For each parameter $\theta_i$, $\delta_i$ denotes the finite-difference step size and $\mathbf{e}_i$ the corresponding canonical basis vector:
\begin{equation}
	\frac{\partial \varCost}{\partial \theta_i} \approx
	\frac{\varCost(\varParamVec + \delta_i \mathbf{e}_i) -
		\varCost(\varParamVec)}{{\delta_i}}.
	\label{eq:fd}
\end{equation}
Evaluating this gradient therefore requires one additional simulation per parameter at each estimation step.

\subsubsection{Adaptive Sliding-Window Estimation}

Windows of length $\varWinLen$ advance along the trajectory by a fixed stride.
For each window, the heightmaps $\varHF$ and $\varHB$ are restricted to the valid-measurement cell set $\varWinCells$ of~\eqref{eq:cost}; the cost and gradient~\eqref{eq:fd} are then evaluated on $\varWinCells$ only, and parameters are updated via gradient descent with per-parameter learning rates $\alpha_i$ and bounds $\theta_i^{\min}$, $\theta_i^{\max}$:
\begin{equation}
	\theta_i \leftarrow \operatorname{clip}\!\left(
	\theta_i - \alpha_i\,\frac{\partial \varCost}{\partial \theta_i},\;
	\theta_i^{\min},\;\theta_i^{\max}
	\right),
	\label{eq:update}
\end{equation}
where $\operatorname{clip}(\cdot,\,a,\,b)$ clamps to $[a,\,b]$.
Critically, each window is initialized from the unmodified, measured $\varHF$ rather than from a previous window's simulated output, so estimation errors do not accumulate along the trajectory.

\section{EXPERIMENTAL SETUP} \label{sec:experiment}

\begin{figure*}[t]
	\centering
	\includegraphics[width=\textwidth]{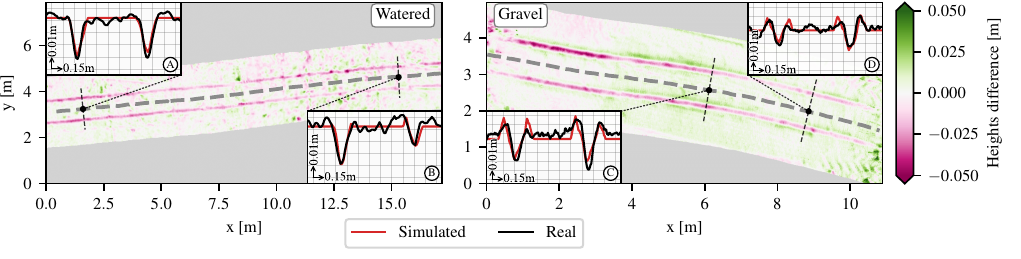}
	\caption{
		Height difference maps $\varHDelta$ on \emph{Watered} (left, windows A and B) and \emph{Gravel} (right, windows C and D).
		Pink indicates soil excavation, green indicates soil deposition.
		Inset cross-sections compare the measured height change to the corresponding simulated prediction across the two wheel tracks.
	}
	\label{fig:heightmap_diff}
\end{figure*}

Experiments were conducted using a four-wheel, dual-steering-axle mobile robotic platform, shown in \autoref{fig:rut}, designed for autonomous navigation in agricultural and unstructured outdoor environments.
The platform has a track width of \SI{1.0}{m}, a wheelbase of \SI{1.38}{m}, and a mass of approximately \SI{700}{\kilogram}.
It is actuated by electric motors on all four wheels, each wheel about \SI{0.14}{m} wide.
Perception is provided by two Ouster OS1-32 lidar sensors (32 beams), mounted at the front and back of the chassis.
This leaves minimal overlap between their fields of view, which rules out conventional target-based or motion-based extrinsic calibration.
We address this with a dedicated procedure, described below.

The robot maintained a constant speed of \SI{1}{\meter\per\second} throughout all trajectories considered in this study.
The relaxation kernel of \autoref{subsec:implementation} uses $\varepsilon = 4\,\Delta x^{-2}$ and a convergence tolerance $\tau = 0.01$ in all experiments reported below.
The \ac{ROI} corridor of \autoref{sec:pipeline} has width $\varPathWidth = \SI{3}{m}$.
In the experiments reported below, the stride is set to \SI{0.4}{m}, shorter than the window length $\varWinLen = \SI{0.8}{m}$, so that consecutive estimation windows overlap.
The estimated parameters are constrained to physically admissible ranges: $\varSinkOne, \varSinkTwo \in [0,\,\SI{0.30}{m}]$, $\varCompRatio \in [0,\,2]$, and $\varAngleRepose \in [\ang{2},\,\ang{90}]$.
The lower bound on $\varAngleRepose$ is \ang{2} rather than \ang{0}, since $\varAngleRepose = \ang{0}$ makes the relaxation step of \autoref{subsec:implementation} numerically unstable.

To calibrate the extrinsics despite this lack of overlap, a complete point cloud of a controlled indoor facility was acquired with a FARO Focus laser scanner, yielding a dense, millimeter-accurate reference map $\mathcal{M}_{\mathrm{calib}}$.
The robotic platform was then driven into the same environment with its lidars rigidly fixed.
For each sensor, \ac{ICP} is run against $\mathcal{M}_{\mathrm{calib}}$ to obtain its extrinsic transform, seeded by a coarse manual measurement.
Composing the two transforms gives the transform between the lidars, used to fuse their point clouds for all subsequent processing.

A separate calibration was performed to establish the pipeline's minimal measurement noise.
The robot was driven along a heavily compacted, non-deformable road within the same structured environment, providing near-ideal \ac{ICP} conditions.
Running the pipeline of \autoref{fig:pipeline} on this path yields a displacement map $\varHDelta$ whose cell values follow a Gaussian distribution with a mean of \SI{1.3}{mm} and a standard deviation of \SI{3.4}{mm}.
Since no genuine soil deformation was induced by construction, the \SI{1.3}{mm} mean reflects a residual registration bias and the \SI{3.4}{mm} standard deviation is taken as the pipeline's minimal achievable measurement noise.
We use this platform and calibration to evaluate the proposed model in the following section.

\section{RESULTS} \label{sec:results}

\begin{figure}[t]
	\centering
	\includegraphics[width=\columnwidth]{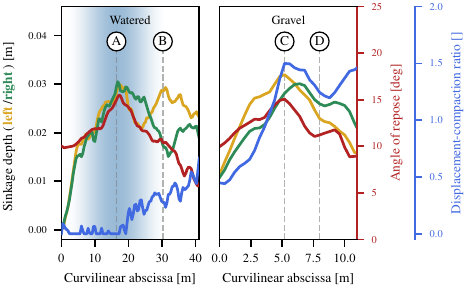}
	\caption{
		Evolution of the identified sinkage depths for the left and right wheels ($\varSinkOne$ and $\varSinkTwo$), the displacement-compaction ratio $\varCompRatio$, and the angle of repose $\varAngleRepose$ over successive estimation windows, for the two field drives.
		Left: \emph{Watered}.
		Right: \emph{Gravel}.
		The markers A, B, C, D correspond to the windows illustrated in \autoref{fig:heightmap_diff}.
	}
	\label{fig:param_convergence}
\end{figure}

The parameter estimation pipeline was evaluated on two continuous field drives, spanning two soil-preparation conditions referred to here as \emph{Watered} and \emph{Gravel}.
On \emph{Watered}, the trajectory begins on a plot prepared with a cultivator and vibro-cultivator, producing a flat, fine-aggregate seedbed, then irrigated across its width with a spatial profile approximated as Gaussian along the direction of travel, to a peak intensity roughly equivalent to \SI{0.020}{m} of rainfall.
On \emph{Gravel}, the vehicle traverses a gravel surface that repeated prior traffic has packed to a dense state.

This section first examines the evolution of the soil state and associated parameters along the trajectories. It then evaluates the contribution of individual model components through an ablation study, followed by an analysis of the accuracy and computational cost as a function of the cell size.

\subsection{Qualitative Analysis}
\autoref{fig:param_convergence} reports the identified parameters over successive estimation windows for both environments, with four representative windows (A--D) used for the analysis. The corresponding difference maps in \autoref{fig:heightmap_diff} focus on the trajectory segment containing these windows, with insets comparing measured and simulated cross-sections perpendicular to the direction of travel.

On \emph{Watered}, the sinkage depth and angle of repose exhibit a bell-shaped evolution around the watering location, consistent with the irrigation profile and the associated change in soil properties.
At point A, both tracks reach similar sinkage depths and the simulated profile closely matches the measured one, while $\varCompRatio$ remains near $0$, consistent with the absence of berms. 
At point B, the sinkage depths diverge, reproducing the observed asymmetric profile, while the increase in $\varCompRatio$ corresponds to a small berm beside the rut. 
This asymmetry reflects local variations in soil resistance arising from differences in composition, water content, and previous traffic.
On \emph{Gravel}, $\varCompRatio$ increases from approximately $0.45$ to $1.5$, with values above $1$ indicating decompaction of the granular material.
This transition is reflected in the cross-sections C and D, where deposition becomes increasingly apparent. 
The angle of repose $\varAngleRepose$ evolves concurrently with the deposition. 
At point C, the parameters and simulated profile are only partially converged, whereas at D, the simulated and measured profiles closely match. 

As such, the identified parameters converge toward regimes consistent with the respective terrain conditions: on \emph{Watered}, sinkage depth and angle of repose follow the irrigation profile while $\varCompRatio$ remains near zero; on \emph{Gravel}, $\varCompRatio$ increases beyond one and the angle of repose evolves with the observed deposition, reflecting the greater propensity of granular materials to displace and decompact compared with agricultural soil.

\subsection{Ablation Analysis}

Two modeling choices in the pipeline are optional refinements whose cheaper alternative saves computation: a single shared sinkage depth removes one forward simulation per gradient step, and capping the erosion relaxation instead of running it to convergence removes iterations per simulation.

\begin{table}[t]
	\centering
	\footnotesize
	\caption{Ablation study: residual and prediction cost ($\times
		10^{-3}\,\mathrm{m^3\,m^{-2}}$) for different model variants.}
	\label{tab:ablation}
	\setlength\tabcolsep{2pt}
	\begin{tabularx}{\columnwidth}{Xrrrrrr}
		\toprule
		                                                & \multicolumn{3}{c}{Residual} & \multicolumn{3}{c}{Prediction}                                                                                                              \\
		\cmidrule(lr){2-4}\cmidrule(lr){5-7}
		Scenario                                        & mean                         & std                            & min/max                    & mean                    & std                    & min/max                    \\
		\midrule
		\textcolor{gray}{Deformation signal} & \textcolor{gray}{9.07}  & \textcolor{gray}{4.33} & \textcolor{gray}{3.9/25.3} & \multicolumn{3}{c}{\textcolor{gray}{/}} \\
		\textcolor{gray}{Noise floor}        & \textcolor{gray}{5.31}  & \textcolor{gray}{3.73} & \textcolor{gray}{2.3/21.7} & \multicolumn{3}{c}{\textcolor{gray}{/}} \\
		Proposed model                       & 5.12 & 2.92 & 2.7/17.3 & 5.96 & 3.17 & 3.0/17.3 \\
		Merged sinkage                       & 5.21 & 2.87 & 2.8/16.9 & 5.96 & 3.16 & 3.0/17.3 \\
		Capped relaxation                    & 5.39 & 4.00 & 2.8/22.4 & 6.45 & 4.64 & 3.0/23.1 \\
		\bottomrule
	\end{tabularx}
\end{table}

\autoref{tab:ablation} evaluates the two ablated variants using the cost defined in~\eqref{eq:cost}, at $\Delta x{=}\SI{0.02}{m}$.
The \emph{residual} cost compares the simulated and measured back surfaces over the same window used for parameter estimation.
The \emph{prediction} cost instead fixes the parameters estimated for each window and evaluates them one window length ahead, on terrain not observed during estimation.
As references, we compute two additional cases characterizing the magnitude of the deformation signal and the measurement noise floor.
The \emph{deformation signal} row compares the undeformed front surface with the measured back surface near the rut; it is the full change induced by the traversal, and serves as a high reference.
The \emph{noise floor} performs the same comparison on undisturbed soil beside the rut, where the two surfaces should coincide, and therefore characterizes the residual sensing and processing error.
Since these reference cases do not involve model-based prediction, no prediction cost is reported for them.
The proposed model achieves a residual cost matching the noise floor, while its prediction cost remains close to it, indicating that the model accounts for the observed deformation down to the noise level.

Merging the independent left and right sinkage depths ($\varSinkOne$, $\varSinkTwo$) into a single parameter increases the residual cost while leaving the prediction cost essentially unchanged.
The asymmetric response therefore improves reconstruction of the traversed rut but provides little benefit for prediction on unseen terrain, suggesting that the parameter adaptation does not track the spatial variation in sinkage rapidly enough.
A single sinkage depth is consequently preferable when the parameters are used for forward prediction, as it provides comparable predictive accuracy with a simpler model: the two-depth formulation is primarily beneficial for mapping the state of the soil via already observed deformation.

Limiting the erosion relaxation to five iterations increases the residual and prediction costs by roughly \SIrange{5}{8}{\percent} and inflates their worst case from about $17\times10^{-3}$ to about $23\times10^{-3}\,\mathrm{m^3\,m^{-2}}$, while reducing the overall computation time by a factor of three.
As such, iterating to convergence provides improved accuracy at the expense of additional computation, while the bounded iteration count may remain appropriate for applications requiring fast computation, such as \ac{MPC}.

\subsection{Computational Performance}
\label{subsec:comp}

\begin{figure}[b]
	\centering
	\includegraphics[width=\columnwidth]{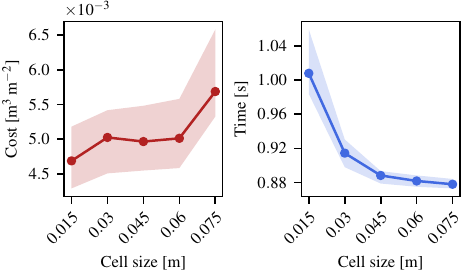}
	\caption{Prediction cost (Left) and computation time (Right) as a
		function of the simulation cell size $\Delta x$. The curves show the
		median over estimation windows, with the interquartile range shaded.}
	\label{fig:cost_time_heatmap}
\end{figure}

This subsection evaluates the effect of the heightmap cell size $\Delta x$ on prediction accuracy and computation time, as reported in \autoref{fig:cost_time_heatmap}. The prediction cost rises from about $4.7\times10^{-3}$ at $\Delta x = \SI{0.015}{m}$ to roughly $5.0\times10^{-3}$ by $\Delta x = \SI{0.03}{m}$, holds near that level up to $\Delta x = \SI{0.06}{m}$, then jumps to $5.7\times10^{-3}$ at $\Delta x = \SI{0.075}{m}$. This degradation is consistent with the spatial scale of the rut, which is approximately one wheel width ($\SI{0.14}{m}$). At $\Delta x = \SI{0.075}{m}$, the rut is represented by only two to three cells, limiting the resolution of its depth profile. This coarse representation also affects the projection of the measured back surface $\varHB$, the single-cell-wide deposition band, and the slope terms in~\eqref{eq:height_update}.

In contrast, computation time decreases from approximately \SI{1.01}{s} to \SI{0.88}{s} as the cell size increases from $\SI{0.015}{m}$ to $\SI{0.075}{m}$, with most of the reduction occurring below $\Delta x = \SI{0.045}{m}$. We hypothesize that, as the number of cells decreases, the GPU kernel launch overhead becomes increasingly significant relative to the computation itself, limiting further reductions in execution time. Consequently, coarsening the map beyond $\Delta x = \SI{0.045}{m}$ provides little computational benefit while continuing to reduce accuracy. The $\Delta x = \SI{0.02}{m}$ resolution used throughout the preceding experiments therefore provides a suitable compromise, retaining high spatial resolution while remaining close to the minimum computation time.

The reported computation times should be regarded as conservative, as the implementation has not been optimized for execution speed and could benefit from dedicated hardware.
At the relatively low speed of agricultural machines, the computation time is compatible with real-time mapping of the soil response and subsequent adaptation of robot behavior.
Further optimization, notably by limiting the maximum number of surface-relaxation iterations, can reduce the computational load and facilitate the integration of the framework into real-time control and planning algorithms.


\section{CONCLUSION} \label{sec:conclusion}
This paper presents an online framework for estimating robot-induced soil deformation from lidar observations.
The proposed approach uses lidar measurements of the soil to estimate a reduced-order parametric model, whose physically interpretable parameters are identified online through an adaptive sliding-window optimization.

Experiments on two different soil conditions demonstrated that the framework can effectively recover the observed soil state and predict its evolution along the robot trajectory.
The identified parameters reflected changes in soil response, while the resulting predictions remained close to the measurement noise level.
As such, the framework provides a compact and continuously updated representation of soil state that can be obtained directly from onboard lidar observations.

The current formulation is limited to quasi-static soil response and does not account for effects that depend on velocity or acceleration.
Future work will extend the method to dynamic conditions and use the predicted soil state to define soil-health cost metrics, such as measures of compaction and overall degradation.
These metrics can then be incorporated into the control and trajectory-planning objectives to balance task performance with the preservation of soil health.
Finally, the framework will be deployed in real agricultural scenarios to assess its practical applicability.
\section{REFERENCES} \label{sec:references}

\printbibliography[heading=none]

\vfill\pagebreak

\end{document}